# A Feature-based Generalizable Prediction Model for Both Perceptual and Abstract Reasoning

Quan Do[1,3], Thomas M. Morin[5,6], Chantal E. Stern[1,2,4], Michael E. Hasselmo[1,2,3]
1. Graduate Program for Neuroscience, Boston University, Boston, MA
2. Department of Psychological and Brain Sciences, Boston University, Boston, MA
3. Center for Systems Neuroscience, Boston University, Boston, MA
4. Cognitive Neuroimaging Center, Boston University, Boston, MA
5. A. A. Martinos Center for Biomedical Imaging, Dept. of Radiology, Mass. General Hospital, Boston, MA
6. Psychology Department, Brandeis University, Waltham, MA

## Abstract

A hallmark of human intelligence is the ability to infer abstract rules from limited experience and apply these rules to unfamiliar situations. This capacity is widely studied in the visual domain using the Raven's Progressive Matrices. Recent advances in deep learning have led to multiple artificial neural network models matching or even surpassing human performance. However, while humans can identify and express the rule underlying these tasks with little to no exposure, contemporary neural networks often rely on massive pattern-based training and cannot express or extrapolate the rule inferred from the task. Furthermore, most Raven's Progressive Matrices or Raven-like tasks used for neural network training used symbolic representations, whereas humans can flexibly switch between symbolic and continuous perceptual representations. In this work, we present an algorithmic approach to rule detection and application using feature detection, affine transformation estimation and search. We applied our model to a simplified Raven's Progressive Matrices task, previously designed for behavioral testing and neuroimaging in humans. The model exhibited one-shot learning and achieved near human-level performance in the symbolic reasoning condition of the simplified task. Furthermore, the model can express the relationships discovered and generate multi-step predictions in accordance with the underlying rule. Finally, the model can reason using continuous patterns. We discuss our results and their relevance to studying abstract reasoning in humans, as well as their implications for improving intelligent machines.

## Introduction

Humans have a remarkable reasoning capacity. In addition to identifying and recognizing similar features to aid categorization, humans can also reason in novel contexts and situations by inferring an abstract rule that is generalizable. This capacity has been studied using the Raven's Progressive Matrices (RPM) (Raven, 1941), one of the most widely used tests of "fluid" intelligence, or the ability to reason about and solve novel problems. In this task, human participants are presented with a 3x3 matrix in which one element, or a component is intentionally left blank. The components in the matrix are related by an underlying rule. The participant must deduce the rules of the matrix to choose the correct component from a given list of available items to fill in the blank. Humans can complete many RPM and RPM-like problems after little or no exposure (Stone & Day, 1981; Vodegel Matzen et al., 1994).

Many AI researchers believe that progress in human-level AI can be made by evaluating their AI systems on the RPM, and the RPM became an important benchmark for testing the reasoning abilities of deep neural networks. Deep neural networks trained to generate texts, classify images, and play video games (Krizhevsky et al., 2017; Mnih et al., 2015; Vaswani et al., 2017) are some of the biggest success stories for AI in the 21[st] century, but they have not achieved human level performance on the original RPM. Surprisingly, a pre-trained Large Language Model GPT-3 (Brown et al., 2020) surpassed human capabilities in a text-based version of the RPM. Results suggest that analogical reasoning may emerge from the sheer quantity of data that the model is trained on - a highly unrealistic scenario for a human reasoner (Webb et al., 2023). The jury is still out on how these large language models are displaying these seemingly emergent capabilities, but deep neural networks embedded with inductive bias on problem solving can also achieve competitive performance on several RPM-inspired datasets (An & Cho, 2020; Barrett et al., 2018; Benny et al., 2021; Hahne et al., 2019; Hu et al., 2022; Jahrens & Martinetz, 2020; Kerg et al., 2022; Malkinski & Mandziuk, 2022; Sinha et al., 2020; Steenbrugge et al., 2018; van Steenkiste et al., 2020; Wu et al., 2020; Zheng et al., 2019). Still, most neural network models are trained to select a correct answer, without providing an explanation for their output, and often rely on biases in the dataset, therefore failing at out-of-distribution generalization (Hersche et al., 2023; Malkinski & Mandziuk, 2022; Sinha et al., 2020; Wang et al., 2020). Furthermore, these neural network approaches focused entirely on the discrete symbolic version of matrix reasoning, even though the Raven's Advanced Progressive Matrices also includes continuous perceptual stimuli (Raven, 1941; Yang & Kunda, 2023). What remains a gap between humans and artificial neural networks is perhaps the ability to flexibly switch between different types of reasoning, perceptual and symbolic, under limited exposure.

It is an open question how humans solve these problems, though modelling efforts include a symbolic production system (Carpenter et al., 1990), geometric analogies (Lovett et al., 2009; Lovett & Forbus, 2017), affine and set transformations (Kunda et al., 2010, 2013; Yang et al., 2022), Bayesian rule induction (Little et al., 2012), role-filler variable bindings via circular convolutions (Rasmussen & Eliasmith, 2011, 2014), and reinforcement learning (Raudies & Hasselmo, 2017). Among these cognitive models, one approach that is largely unexplored in the AI community is affine transformation, even though evidence from psychology and neuroscience (DeShon et al., 1995; Kunda & Goel, 2008; Prabhakaran et al., 1997; Soulières et al., 2009) suggests that human participants frequently used affine transformation to solve RPM tasks. Intuitively, this process would involve inspecting the objects in the matrix, comparing the objects, and mentally transforming the objects and estimating their perceptual similarity. Therefore, computational models in this line of approach (Kunda et al., 2010, 2013; Yang et al., 2022) represent the matrix entries in the RPM by pixel images, apply predefined pixel-operations to, and calculate pixel-level similarities between these images. However, since the human eyes cannot differentiate individual pixels on a computer screen, it's unlikely that humans have the same pixel level representation as these computational models. Additionally, the major downside of these approaches is the computational power required to operate on all pixels in a large image, or the lack of sufficient abstraction that reduces the complexity of problem solving. In fact, evidence from neural imaging studies (Schendan & Stern, 2007, 2008) suggests that humans applied mental rotation to a more abstract representation like object.

Here we further explored the affine transformation approach and introduced a feature-based algorithmic framework for both perceptual and symbolic reasoning. Instead of representing and acting on pixels, explicit symbols, or bound variables, our model works with scale-invariant features detected with the SIFT algorithm (Lowe, 1999), which share several properties in common with the responses of neurons in the IT cortex in primate vision (Ito et al., 1995). These features become the common denominators shared among multiple reasoning demands, perceptual and symbolic. Reasoning in our model involves iteratively estimating and deriving a sequence of affine transformations acting on the detected features via a model fitting procedure called random sample consensus (RANSAC) (Fischler & Bolles, 1981). The sequence is fine-tuned by similarity metrics between the predicted and desired outputs to generate a set of interpretable and generalizable operations that can be used for extrapolation.

We applied our approach to a modified version of the RPM which was designed for a neuroimaging study of both symbolic and perceptual reasoning skills in humans (Morin et al., 2023). Our modelling approach, under the same task structure and condition that human participants experienced, performed near human level on the symbolic conditions and well above chance on the perceptual conditions. The model returned interpretable and generalizable relationships given only a small number of observations, unlike any of the existing deep neural networks models. Furthermore, our work provided a novel interpretation of the human neuroimaging results from Morin et al 2023, namely that the functional reconfiguration of the frontoparietal network during abstract reasoning could reflect an iterative process of searching for a generalizable sequence of transformations needed to capture the underlying rule.

## Methods/Approach
**Task Design.**
In our modelling approach, the artificial agents experience the same task structure that human participants experienced in a modified and simplified version of the RPM (Morin et al., 2023). There are four task conditions in this modified RPM to probe four cognitive abilities: Perceptual Matching, Perceptual Reasoning, Symbolic Matching and Symbolic Reasoning. Each task condition contains 24 unique cue stimuli. Each cue stimulus was displayed four times: twice normally, and twice left/right flipped (reflected over the y-axis), resulting in 384 unique trials total for the task. Along with the presentation of the cue stimuli, two answer choices are displayed on each side of the screen. The agents must make a response (0 for left and 1 for right) to indicate their answers. The four task conditions are interleaved on a trial-by-trial basis, and the agents completed all 384 trials of the task.

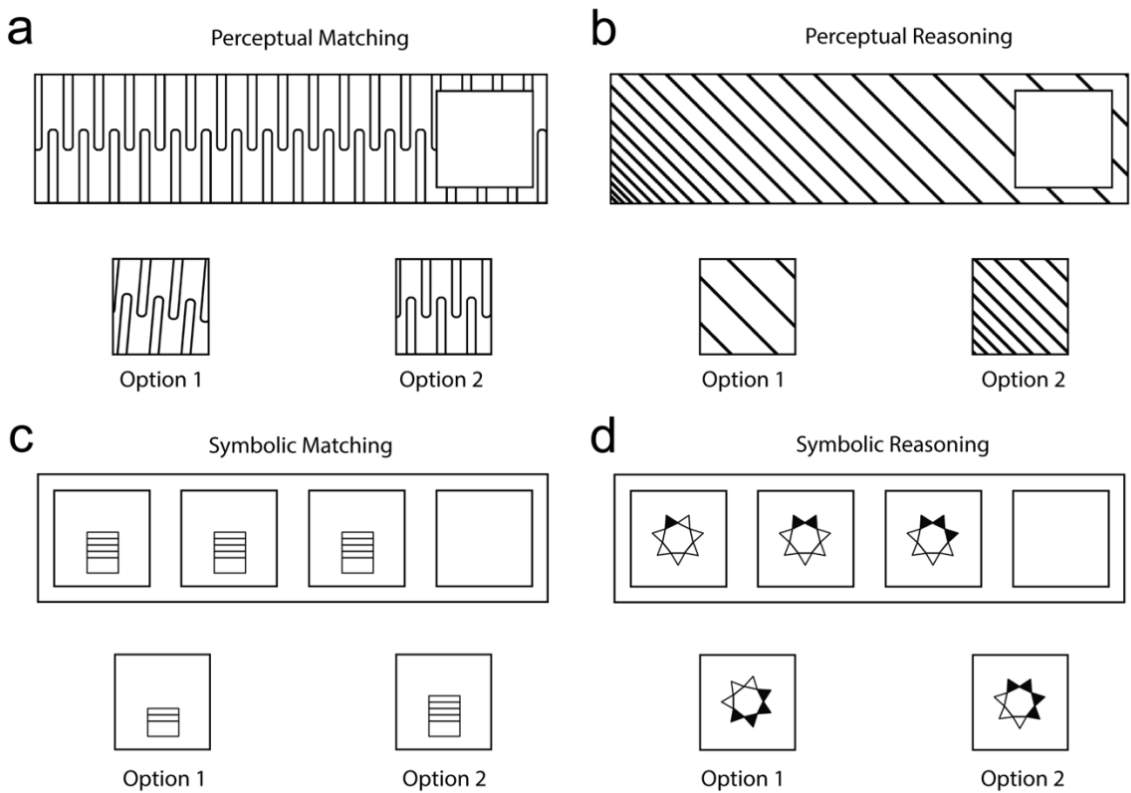

**Figure 1: Task Design**
(a) A condition to test Perceptual Matching ability. A continuous pattern is presented with two answer choices. Agents pick one of the two options to fill in the blank. Option 2 is the correct answer. (b) A condition to test Perceptual Reasoning. Option 1 is the correct answer. (c) A condition to test Symbolic Matching. A discrete set of symbols appears in series with two options to fill in the blank. Option 2 is the correct answer. (d) A condition to test Symbolic Reasoning. Option 2 is the correct answer.

**Trial Identification and Attention Window Detection.**
In the Symbolic Matching and Symbolic Reasoning conditions, the different cues are located inside three horizontally and evenly spaced rectangles. A fourth rectangle, the last one on the right, is left blank, indicating where the answer would be. The two answer choices displayed below the cues on each side of the screen are also presented within rectangles of equal dimensions (**Figure 1c, d**). From the artificial agent's perspective, the rectangles containing the answer choice have the exact same dimensions as the blank rectangle. The artificial agent runs a contour detection procedure on the cue stimulus to get polygons that best approximate the detected contour. By checking for the number of vertices as well as the angles between the edges of the polygons, the artificial agent can reliably find all the cue rectangles as well as the blank rectangle. The number of detected rectangles also tells the artificial agent which trial conditions it is in. To verify that the detected rectangles indeed contain the cues, the artificial agent also compared the dimensions of each cue rectangle with the dimensions of the rectangles containing the choice options, and only kept the ones that matched. The image content inside the detected rectangles is the only piece of information the artificial agent pays attention to. This procedure results in three attention windows, one for each presented cue, labelled cue A, B and C.

In the Perceptual Matching and Perceptual Reasoning conditions, there is only one rectangle inside the cue stimulus that is left blank (**Figure a, b**), indicating where the answer would be for the human participants to fill in. The two answer choices displayed below the cues on each side of the screen are presented within rectangles of equal dimensions (**Figure 1a, b**). The artificial agent in these conditions also runs a contour detection procedure on the cue stimulus to reliably find the blank rectangle and calculate its width and height. The artificial agent can check that the detected rectangle is correct by comparing its dimension to that of the rectangle containing the answer choice. The number of detected rectangles also tells the artificial agent which trial conditions it is in. For the Perceptual Matching and Perceptual Reasoning conditions, only one rectangle should be detected. The artificial agent subsequently divides the cue stimulus into 4 equally sized rectangles that span the height of the original cue stimulus, which we thereby labelled Cue A, B, C, D. Only cue D contains the blank square where the answer would be filled in. The cues A, B, C are the attention windows that the artificial agent focuses on.

**Feature Detection and Matching.**
Once an attention window is defined, the agent looks for SIFT features based on the content inside the window (**Figure 2a**). We also tried ORB (Oriented FAST and Rotated BRIEF) feature detection (Rublee et al., 2011) since it provides a free and efficient alternative to the popular SIFT algorithm. SIFT and ORB are interchangeable for our purposes. We however noticed that SIFT returns more features than ORBs and is better for the Perceptual conditions. ORB can fail to find features in these conditions.

In all four conditions (Symbolic Reasoning, Perceptual Reasoning, Symbolic Matching, Perceptual Matching) of the task, there are three attention windows for the three distinct cues. The detected features from these three windows are compared and matched using a brute force matching procedure, to locate the repeating features across the three cues. A visual feature consisted of a descriptor and a key point (that defines the location on the screen in x-y coordinates). One descriptor for a feature from cue A was compared with all the descriptors of features from cue B, using the Hamming distance, until a best match was found. Then the next descriptor for a feature from cue A was matched to all those in cue B, and so on. After all the matches are determined, only the closer descriptors are kept for quality control. This was done with a nearest neighbor distance ratio of 0.8 (arbitrary choice). Finally, the x-y coordinates of the detected, matched and chosen descriptors between cue A and B, provided by the key points, were then used as input into the Transformation Estimation step.

**Transformation Estimation/RANSAC Outlier Detection and Local Similarity**
In this step, after the matched features or correspondences across A, B and C are detected, we assume that there exists a repeatable sequence of geometric transformations between these features, and we try to estimate the parameters of these transformations as well as the number of steps in the sequence.

Random sample consensus (RANSAC) was used to robustly estimate parameters of the geometric transformation between Cue A and Cue B. This ensures that we are not making errors due to faulty matched features if we solved for the transformation using the full set of detected features.

RANSAC returns the parameters for a geometric transformation that best fits the largest set of features. The geometric transformation can include affine transformation components including rotation, translation, and size changes. The features that are effectively fitted to the transformations are referred to as the 'inliers.' RANSAC also returns the x and y coordinates of the 'outliers' that do not fit the transformations (Bottom of **Figure 2a**). We call the initial inlier set I and the initial outlier set O. We randomly sample 3 inliers from the set I and run RANSAC again on the outliers as well as the 3 recently sampled inliers (minimum of three points are needed to fit a geometric transformation) to detect whether another distinct transformation exists across a different set of point combinations. Rotation and translation only become detectable at this stage when computing transformations on outliers, whereas the previous step had identified either identity or similarity transformations. Upon finding a new transformation, we update the outlier set O with the reduced number of outliers. We continue running this procedure until there are no more outliers (**Figure 2b**). At each iteration, we computed a local similarity index. Since there is a stochastic component to the sampling of inliers at each step, resulting in different transformations being found, we do this step several times (10 times). A local similarity index is computed by applying the local transformation to cue B to predict cue C and calculating the mean squared error (MSE) between the prediction and the desired output. Note that the main difference here is instead of applying a transformation to the detected features, we are applying the transformation to every pixel in the image using an image warping function. We are distorting a part of cue B to match a part of cue C. We only kept the local transformation that results in the lowest MSE.

In summary, given cue A and B, we followed a random sampling and greedy strategy to look for the sequence of actions or transformations that returns the best local prediction of cue C (a local transformation transforms a part of cue B to match a part of C) at each step. We note that there are other strategies to sample and update the set of inliers and outliers that are yet to be explored. It's an open question what strategy a human observer would follow.

**Image Combination, Thresholding and Global Similarity.**
To verify that the transformed output matches the desired output, we apply all the transformations in the detected action sequence to the input, add up the resulting outputs, and threshold the final output. Input and output refer to one pair of images being tested, in this case cue A and B, respectively. At each local transformation step, a non-biased coin is tossed to decide whether the round up or down the outputs generated by the transformations, which effectively allow us to add or subtract image components (**Figure 2b**, Bottom of **Figure 3a**). Essentially, given a thresholding value randomly chosen from a preselected list (half the max pixel value, two-third the max pixel value, or one-third the max pixel value), overlapping components are enhanced, resulting in addition, whereas non-overlapping components are removed, resulting in subtraction. There might be a better way to combine image components, but we assume that the default strategy is to transform at each step, add all the previously transformed outputs and then randomly threshold the outputs. We compute a similarity index between the transformed output to the desired output. We again used the mean squared error (MSE) between two images as a measure of global similarity. A lower mean squared error indicates a better fit.

We ran the greedy local search procedure described above multiple times (2 or 3 times), each time with different thresholding values randomly sampled from the preselected values, computed the global similarity index, and selected the search thread that resulted in the lowest MSE.

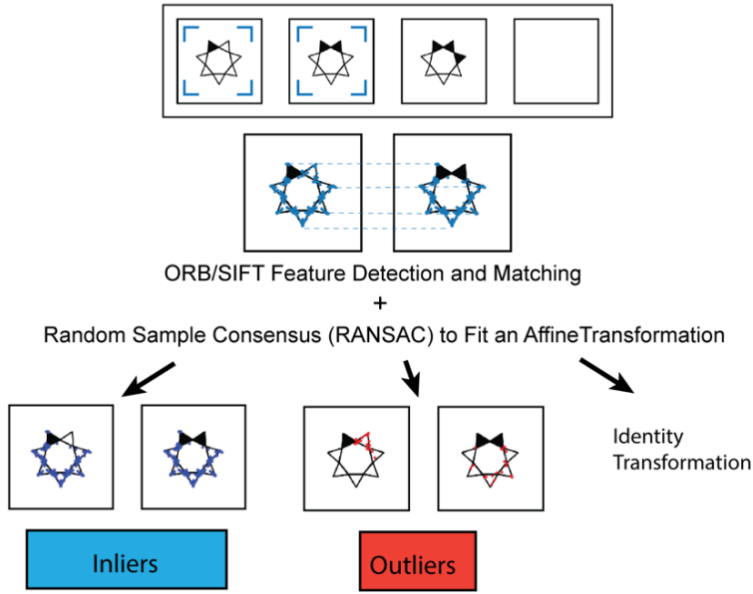

Thresholding is important because at each step, we are distorting a part of a cue to match it to another part of another cue, and we want to be able to subtract the parts that aren't relevant after the distortion process or highlight the parts that might be important. This process would effectively let us manipulate and combine parts of an image to transform it to a different image. We apply the sequence of steps from the selected search thread to cue C to predict the answer in the blank square.

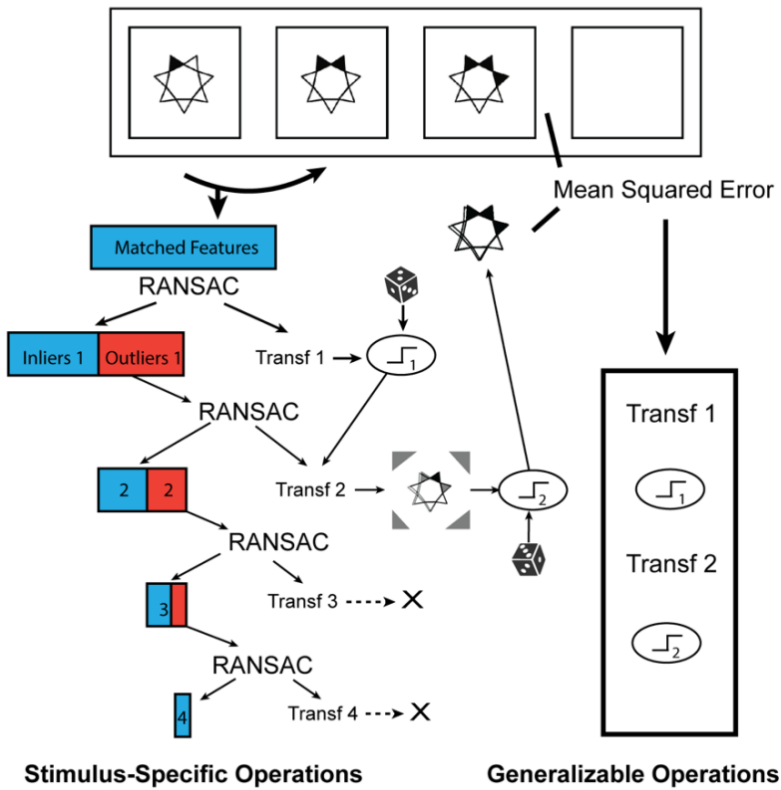

**Figure 2: Operations Fitting and Searching Procedure.** a) Matched ORB/SIFT features between the source and target image were used to fit an affine transformation with RANSAC. Example of an identity transformation returned by RANSAC. The outliers (red dots) are points that cannot be fitted by the identity transformation while the inliers (blue dots) are points that are identical between the source and target. b) After the first run of RANSAC, the outliers can be fitted to another RANSAC procedure, and so on, to generate a list of affine transformations that will fully transform all the matched features from the target to the source for a specific set of input-output pair. Each transformation is followed by a thresholding function with a threshold value randomly drawn from a list of prechosen thresholds. The transformation and thresholding function are then iteratively applied to the target to predict the next stimulus in the sequence. Mean Squared Errors between the predicted stimulus and the actual stimulus determines whether a particular transformation should be kept, and whether the chosen thresholding value would lead to a better output. The model arrives at a final list of operations that can act on each of the input images in the sequence.

### Results:
**The model found generalizable operations from stimulus-specific operations, allowing for interpolation and extrapolation under different rules.**

The model after the fitting and searching procedures can arrive at a list of operations. We formulate the structure of these operations (**Figure 3a**) to describe how these are iteratively applied to the input. Essentially, after a transformation, an output is added to the previous output and a thresholding function is applied to either highlight the important parts or remove the extraneous parts of the image.

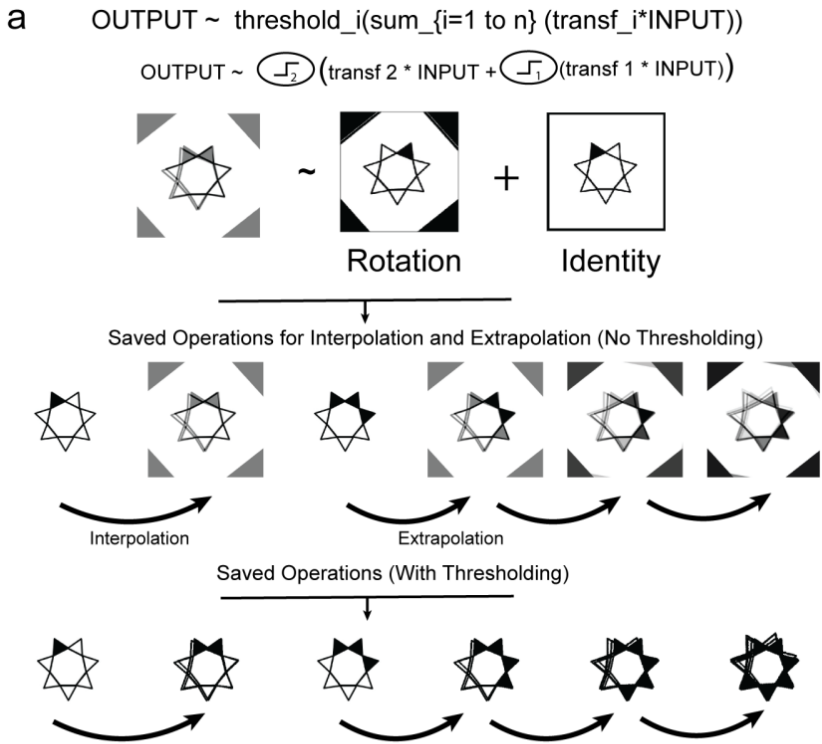
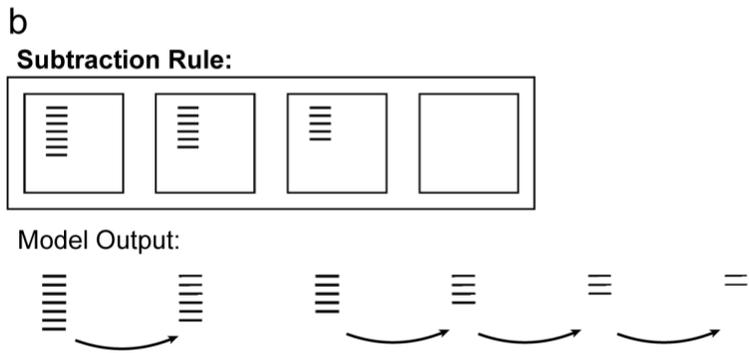
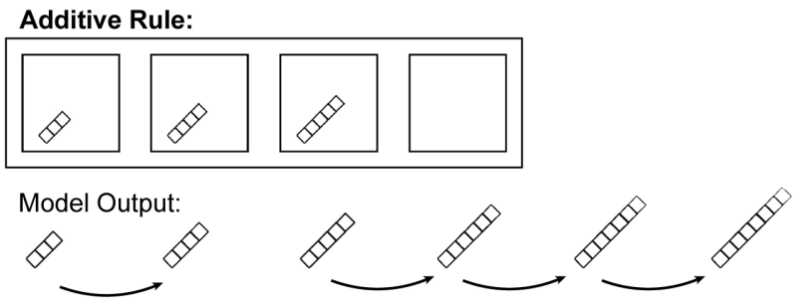

**Figure 3: Model finds generalizable solutions that allow for extrapolation.** a) A general formulation for the operations that can be discovered by the model. The formulation can be understood to a composition of multiple affine transformations with the output going to different thresholding functions at each step. An example of a rotation and identity operation being combined and applied to an input image. Applying the combined operations multiple times enables multi-step predictions that are in accordance with the underlying rule. Adding the thresholding functions allow for removal of the background. b) Examples of the model working on stimulus sets governed by subtraction rule and an additive rule. While the affine transformation acted on the entire image, the thresholding functions allow for robust subtraction and addition of parts of the image. Specifically, overlapping parts are enhanced resulting in addition whereas non-overlapping parts are removed, resulting in subtraction.

For instance, a rotation can be combined with an identity transformation to color an additional part of the hexagram (**Figure 3a**). The part of the hexagram that is essential for highlighting successful rule application, can then be highlighted after applying the thresholding function (**Figure 3a**). With the combination of the transformation and thresholding function, the model can therefore manipulate parts of the image instead of acting on the entire image, allowing for a more robust image manipulation scheme. Two examples further highlight how the model can subtract and add parts to an image (**Figure 3b**). Finally, after discovering these operations, the model can apply them iteratively to an image and its subsequent outputs to generate predictions multiple steps into the future (**Figure 3a, b**).

**The same model can be applied to continuous stimuli, demonstrating complex pattern completion and prediction.**

Humans can generalize using continuous perceptual stimuli. Here we demonstrated that our model can be applied to continuous patterns in addition to discrete symbols. We made no significant changes to the model as described in Figure 2. The only change is that instead of looking at the three attention windows as defined by the three squares in the Symbolic Reasoning Condition (**Figure 1d**), the model divides the continuous patterns into 4 equal slides, and operates on the three slides containing the patterns, while trying to predict the slide containing the blank square (See Methods). Going through the same fitting and searching procedure as described in Figure 2, the model can generate complex patterns to fill in the blank and extrapolate the patterns beyond what was given (**Figure 4**).

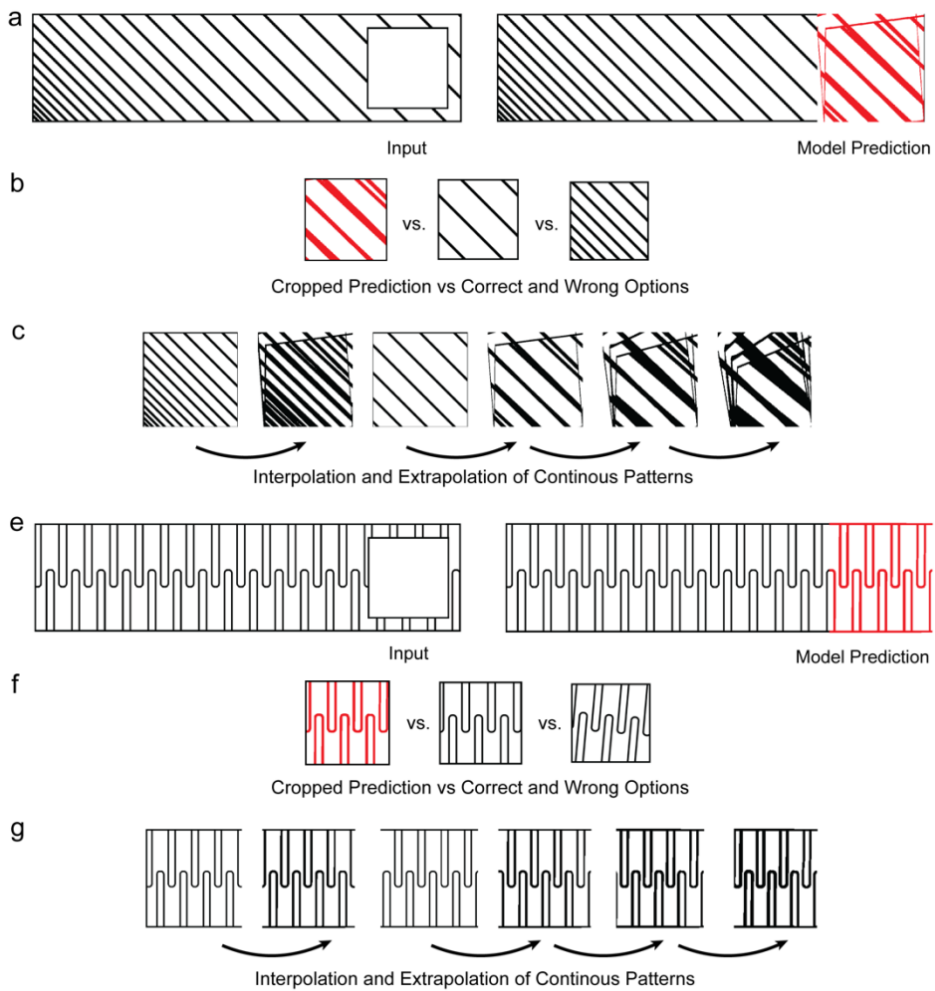

**Figure 4: Model works on continuous perceptual stimuli to perform pattern completion and extrapolation.** Two examples of the model generating continuous patterns to fill in the blank and to extrapolate, beyond what was given. a, e) Given a pattern with a blank square, the model can predict the missing patterns. b, f) The full prediction can be cropped to the size of the blank square and compared to the different answer options. c, g) The model demonstrates interpolation and extrapolation with continuous patterns.

The model prediction is noisy and declines in quality over multiple future steps, but for the task of filling in the blank, the first prediction from the model is sufficient, and can be directly compared to the list of available answer choices to make the correct decision.

**The model achieved comparable performance to humans on most of the task conditions.**

We ran the model on the 384 trials that humans are given in the modified Raven task (Morin et al, 2023). The task has 4 conditions that tests different perceptual and symbolic reasoning skill (See Methods). The Symbolic Rule and Matching condition tests reasoning with discrete symbols while the Perceptual Rule and Matching conditions test reasoning ability on continuous patterns. On a single best run, the model solved 86/96 trials in the Symbolic Reasoning condition, 96/96 trials in the Symbolic Matching condition, 78/96 trials in the Perceptual Matching condition and 63/96 trials in the Perceptual Reasoning condition. Since there is a stochastic component to the model, we ran the model 5 times on the full task to simulate 5 artificial agents (arbitrary choice) and to compare with the data collected from 27 human participants tested on the task (behavior data from Morin et al, 2023, excluding subjects (n=1) who failed to achieve at least 80% accuracy on each of the four conditions). We plotted the human data and the model's performance side by side, grouped by the 4 conditions of the task (**Figure 5a**).

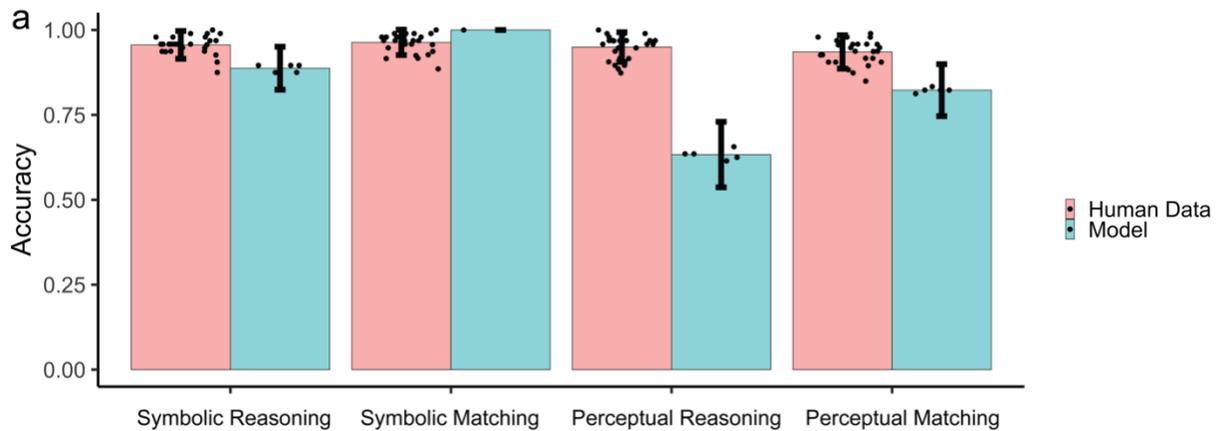

**Figure 5: Model versus humans on different task conditions.** a) Comparing the performance of the model with human participants performing the task. The model performs well above chance and reaches close to human level in several task conditions. Error bar indicates trial-to-trial variability (384 trials). Scatter dots indicates individual variability (27 humans and 5 artificial agents).

We noted that the models performed well on the Symbolic Reasoning (88.75% accuracy, std. = 31.63%) and Symbolic Matching (100% accuracy, std. = 0%) condition. Humans achieved a 95.6% (std. = 20.5%) and 96.36% (std. = 18.73) accuracy rates on these conditions, respectively. The model also works with the continuous texture task conditions and can achieve an 82.29% accuracy rate (std. = 38.21%) on the Perceptual Matching condition, compared to human level of 93.54% (std. = 24.58%). However, the model struggled on the Perceptual Reasoning condition when compared to human performance (mean = 94.96%, std. = 21.86%), despite performing above chance (63.33% accuracy rate, std = 48.24%).

We further explored the relationship between the model and humans by comparing the number of scale-invariant features detected and matched by the model (**Figure 2a**) with the performance of both the model and humans on different task conditions. We fitted linear models and found that only in the Perceptual Reasoning condition, both humans (intercept=0.94, slope = 7.344e-5, p-value = 0.0046, R-squared = 0.003) and the model (intercept=0.59, slope = 0.0004, p-value = 0.0023, R-squared = 0.019) perform better when there are more scale-invariant features found. This is somewhat surprising since one would suspect that humans might perform better on visual inputs with less features. The model however would predict that more features mean better chance of finding and fitting better affine transformations with RANSAC. Our findings suggest that this prediction might hold true for humans.

**Discussion**

In this work, we introduced a generalizable prediction model that can reason with both continuous patterns and discrete symbols. Given a series of three observations, continuous or discrete, our model returns an interpretable and generalizable sequence of operations that can be used to predict multiple future items in the series. The operations that the model discovered are reflective of the multiple underlying relationships that governed the series of observations. In a sense, the model has begun to capture some element of inductive reasoning, or the process of drawing general conclusions from a few specific observations, which is what the original RPM task was designed to test (Kunda et al, 2023). We discuss the applicability of our model as well as some future directions.

**Simplified Raven Task and Applicability to other Raven-like tasks.**
Most RPM and RPM-like tasks usually have a 3x3 matrix structure whereas the modified Raven task we are using has a 1-dimensional structure. This raises the question of how applicable our model is to the traditional RPM as well as other Raven-like tasks. We argue that the core of the RPM, problem solving requires inferring the relationship between the different matrix elements. The spatial structure of a task is arbitrary. The spatial structure may guide the rule inference process, but it is not crucial that the relations being inferred must strictly obey the spatial arrangement. Therefore, our model, though tested on a fundamental 1-dimensional series structure, is directly applicable to other Raven-like tasks, since its core process is relational inference. However, it is interesting to note that in most RPM and RPM-like tasks, including the task we are using, the relations being inferred often follow a left-to-right, top-to-bottom spatial directions. This spatial bias is hard-coded into our model. In a 3x3 task where the blank matrix element is in the top left corner instead of the bottom right, our model would fail to generalize, whereas a human participant would most likely still be able to figure out the rule. There would need to be an additional process in the model that could flexibly infer how the spatial structure of the task relate to the relations being inferred. We have yet to implement this process since it is not necessary for the task we are using, but we think it involves testing our relation inference process along different vectors pointing in different directions (left to right, right to left, top to bottom, bottom to top, etc.) by trial-and-error.

Another key difference between the modified RPM task we are using, and the full RPM task is that ours only includes "sequence" items. The full task also includes items that require an "addition/subtraction" or "set membership" operation (see Rasmussen & Eliasmith 2014 for a good conceptualization of this). These are operations that could be included in the model that we proposed. However, we suspect that we would need explicit symbols, or representations more abstract than scale-invariant features to work with these operations.

**Relation Inference with RANSAC.**
Perhaps the most important part of our model is the Random Sample Consensus (RANSAC) approach to fitting an affine transformation. Typically, the standard way to estimate an affine transformation is by using linear least squares (Zikan, 1991). This is essentially finding a best fit linear function to the data points. However, if we were to be fitting a function using a linear regression to infer a relationship, and if there were multiple relationships between the datapoints, this fitted function would be skewed away from the true relationships. The function fitting process would output an average relationship for all the datapoints. This is not desirable for a model that wants to capture all the underlying relationships. RANSAC however splits the data into inliers and outliers, and the fitted function is fitted only to the inliers, giving us one specific relationship that governs the inlier sets. We can then move on to the outlier set to infer the next relationship. This is beneficial in all the Raven tasks because the different parts within a matrix element have different relationships with their corresponding parts in another matrix element. The rule usually involves identifying and inferring these distinct relationships and applying them accordingly.

**Part-Whole Image Manipulation with Adaptive Thresholding.**
Once all the individual relationships are inferred, we need to be able to identify and manipulate the different parts of the matrix element. This is a challenging step, and we came up with an ad-hoc solution where we still apply the transformation to the entire matrix element instead of searching for individual parts. We rely on the thresholding function to add or subtract parts of the image. Overlapping parts are typically important and are retained with the thresholding function whereas non-overlapping parts are typically deleted. This is however not an ideal solution since in a task with complex patterns, we notice that the predicted pattern often gets corrupted, therefore hindering the model performance. This was noticeable especially in the Perceptual Reasoning condition, where our model underperformed. Ideally, we should crop different regions of the image and apply the appropriate transformation to only the parts that were cropped. To do this, we however would need to be able to identify which parts of the image correspond to which feature sets. Object or shape detection, as well as the understanding of part-whole relations within an image would be beneficial here to constraint the mapping process.

**Continuous Patterns, Scale-Invariant Features and Discrete Symbols.**
Our model works on both continuous patterns and discrete symbols despite using scale-invariant features as the underlying representations. It is not difficult to imagine that scale-invariant feature is a bridge that links our continuous perceptual experience of the world to our symbolic representations. This point is illustrated in the field of computer vision, where the standard workflow involves detecting scale invariant features from a complex visual scene (a continuous perceptual experience) to identify a category or concept (a symbolic representation). Here we show that for most of the reasoning demands in the simplified Raven task, scale invariant feature is the right level of abstraction. Humans can flexibly switch between the different levels of abstraction to suit the reasoning demands, but this is beyond the scope of our work described here.

**Relationship to Functional Reconfiguration of Frontoparietal and Visual Networks.**
Using network analysis on fMRI data, work from our lab (Morin et al 2023) found a stable community structure among frontoparietal and visual brain regions that formed during the simplified Raven task. This community was maintained across all four task conditions, namely Perceptual Reasoning, Perceptual Matching, Symbolic Reasoning, and Symbolic Matching. We postulated that the formation of a strong frontoparietal-visual community facilitates the integration of visuospatial information. This is consistent with our model that works on all task conditions, and consistent with the process of fitting affine transformation to scale invariant features. Furthermore, we found that the frontoparietal cortex was significantly more active for the reasoning conditions compared to the matching conditions. This is also consistent with the model since in the matching conditions, typically only an identity relation is sufficient for the discrete symbols, and sometimes a simple translation operation for the continuous patterns. Humans also showed increased activity in inferior temporal cortex on the perceptual reasoning condition compared to the perceptual matching condition (and overall, on the perceptual conditions compared to the symbolic conditions). This is consistent with the idea that there are many more relevant scale-invariant features that need to be detected in the perceptual conditions, especially in the perceptual reasoning condition. It also supports our finding of a linear relationship between human performance in the perceptual reasoning condition and the number of scale-invariant features detected by our proposed model. One finding in the fMRI study that was not consistent with our model is the observation that the cognitive control network has greater activation during the symbolic reasoning condition in comparison to the perceptual reasoning condition. This may suggest that human participants are using more abstract representations like shape and object in the symbolic reasoning condition, instead of the scale invariant features employed by our model. As previously discussed, our model would greatly benefit from the incorporation of an additional process that cluster and categorize features into more abstract representations.

**Algorithms, Out-of-Distribution Generalization, and Neural Networks.**
Humans have a remarkable ability to generalize and extrapolate under various contexts even with a limited number of observations. In this paper we present an algorithmic framework that learns in one shot, generalizes, and extrapolates across a wide range of task demands. It's hardly surprising that an algorithm hand-crafted to a specific task can have good task performance and generalize to various instances of the task. However, to solve a novel task with varying level of complexity (P vs NP, for instance) would require significant redesigning of the algorithm. The real challenge is in designing an algorithm that can rewrite itself to adapt to the changing task demands. Our algorithmic framework illustrates how simple operations can be flexibly composed to perform different tasks. We argue that the tasks we are using, while simple to perform, vary in their reasoning demands, as shown by the activations of different brain networks under different task conditions by Morin et al, 2023.

One limitation of our work, and the limitation of algorithmic design in general, is the data preprocessing and feature selection that constrains the algorithm to work only on our chosen task domain (image sequence). Deep Neural Networks, only the other hand, have been successful in solving tasks of many domains including but not limited to texts, images, discrete actions, and proprioceptive inputs (Reed et al., 2022). It's natural to then wonder whether neural networks can mimic algorithms, to get the best of both worlds. This is the major focus of an emerging field called Neural Algorithmic Reasoning (Veličković & Blundell, 2021). Unfortunately, the out-of-distribution generalization capability of deep neural networks is lacking and not well-understood. For example, going back to Raven-inspired tasks, the state-of-the-art performance of neural networks on the extrapolation regime of the Procedurally Generated Matrices (Barrett et al., 2018), in which test problems contain feature values outside the range of those observed in the training set, is currently at 25.9% (Malkinski & Mandziuk, 2022; Sinha et al., 2020; Wang et al., 2020). It's therefore unclear what classes of algorithms a neural network can mimic, and whether neural networks can discover novel algorithms beyond the training set, though progress is being made (Bevilacqua et al., 2023; Ibarz et al., 2022; Xu et al., 2020, 2021).

Out-of-distribution generalization is a daunting yet exciting challenge. In future work, a major focus of ours will be to learn from how human reasons, build upon our algorithmic framework, and create neural circuits that generalize. Ultimately, human problem solving, regardless of how complex, is reducible to biological neural networks performing computations.

**Acknowledgements**
This work is supported by the Office of Naval Research ONR MURI N00014-19-1-2571, ONR MURI N00014-16-1-2832, NSF 1625552, ONR DURIP N00014-17-1-2304, and Kilachand Fund Award.